\documentclass{article} 
\usepackage[nonatbib,final]{arxiv_nips_2016}
\usepackage{hyperref}
\usepackage{url}
\usepackage{amsmath,amssymb}
\usepackage{algorithm}
\usepackage{algpseudocode}
\usepackage{graphicx}
\usepackage{subfigure} 
\usepackage{wrapfig}
\usepackage{array}
\usepackage{multicol}
\usepackage{caption}  
\DeclareCaptionType{noticebox}
\usepackage{tikz}
\usepackage{mathtools}

\usetikzlibrary{arrows, decorations.markings}
      \tikzstyle{vecArrow} = [thick, decoration={markings,mark=at position
      1 with {\arrow[semithick]{open triangle 60}}},
      double distance=1.4pt, shorten >= 5.5pt,
      preaction = {decorate},
      postaction = {draw,line width=1.4pt, white,shorten >= 4.5pt}]
     \tikzstyle{innerWhite} = [semithick, white,line width=1.4pt, shorten >= 4.5pt]
\usetikzlibrary{positioning}

\title{Tracking Objects with Higher Order Interactions using Delayed Column Generation}
\author{
Shaofei Wang\\
\texttt{sfwang0928@gmail.com}
\And
Steffen Wolf\\
Heidelberg University, Germany\\
\texttt{steffen.wolf@iwr.uni-heidelberg.de} 
\And
Charless C. Fowlkes\\
University of California, Irvine\\
\texttt{fowlkes@ics.uci.edu} 
\And
Julian Yarkony\\
Experian Data Lab, San Diego, CA \\
\texttt{julian.e.yarkony@gmail.com} 
}

\begin{document}

\maketitle

\begin{abstract}
We study the problem of multi-target tracking and data association in video.
We formulate this in terms of selecting a subset of high-quality tracks subject
to the constraint that no pair of selected tracks is associated with a common
detection (of an object). This objective is equivalent to the classic NP-hard
problem of finding a maximum-weight set packing (MWSP) where tracks correspond to sets and is made further
difficult since the number of candidate tracks grows exponentially in the
number of detections.  We present a relaxation of this combinatorial problem
that uses a column generation formulation where the pricing problem is solved
via dynamic programming to efficiently explore the space of tracks.  We employ
row generation to tighten the bound in such a way as to preserve efficient
inference in the pricing problem.  We show the practical utility of this
algorithm for tracking problems in natural and biological video datasets.  
\footnote{This work was supported in part by NSF grant IIS-1253538}
\end{abstract}
\section{Introduction}

Multi-target tracking in video is often formulated from the perspective of
grouping disjoint sets of candidate detections into ``tracks'' whose underlying
trajectories can be estimated using traditional single-target methods such as
Kalman filtering.  There is a well developed literature on methods for
exploring this combinatorial space of possible data associations in order 
to find collections of low-cost, disjoint tracks.

We highlight three approaches closely related to our method. Approaches based on
reduction to minimum-cost network flow \cite{flownet} map tracks to unit flows
pushed through a network whose edge costs encode track quality. This elegant
construction utilizes edge capacity constraints to enforce disjoint tracks
and allows for exact, polynomial-time inference.  However, this formulation is
quite limited in integrating joint statistics over multiple detections assigned
to a track.  In particular, it is constrained to first-order dynamics in which
the cost of a detection being associated with a given track depends only on 
the immediately neighboring detections.

Multiple Hypothesis Tracking
\cite{blackman2004multiple,poore1993lagrangian,kim2015multiple} attempts to
overcome these limitations by grouping short sequences of detections into a set
of hypothesized tracks that can be evaluated and pruned in an online manner.
This trades efficiency and global exactness of min-cost flow trackers for
additional modeling power. For example, the cost of a track may be computed
using, e.g.  spline-based fitting of trajectories and instance specific
appearance models.  However, such methods face a combinatorial problem of
assembling compatible sets of tracklets which is usually tackled using greedy
approximations.

Our method is most closely related to the Lagrangian relaxation method of
\cite{lagtrack}, which attempts to capture the speed and guarantees min-cost
flow tracking while maintaining the modeling advantages of tracklet approaches.
A large number of short sequences of detections (subtracks) are generated, each
of which is associated with a cost.  The set of subtracks form the basis from
which tracks are constructed. The corresponding  optimization is attacked via
sub-gradient optimization of the Lagrangian corresponding to the constrained
objective.

Inspired by \cite{sontag}, we attack the problem of reasoning over subtrack
assembly constraints using column/row generation
\cite{cuttingstock,barnprice,HPlanarCC} to provide faster inference with
tighter bounds than \cite{lagtrack}. This paper is organized as follows.  In
Section \ref{lpcliquePrelim} we formulate tracking as optimization of a linear
programming (LP) relaxation equivalent to \cite{lagtrack}.  In Section
\ref{betterLP} we demonstrate a simple case in which the LP relaxation is loose
and demonstrate how to tighten the bound.  In Section \ref{tightlpopt} we
formulate optimization over the tighter bound  and discuss inference using
column and row generation.  In Section \ref{Exper} we demonstrate the
effectiveness of our approach on pedestrian tracking and biological image
data benchmarks.

\section {Constraint Relaxation for Multi-target Tracking}
\label{lpcliquePrelim}
\subsection{Feasible Trackings}
Given as input a set of candidate detections $\mathcal{D}$, each with a
specified space-time location, our goal is to identify a collection of tracks
that describe the trajectories of objects through a scene and the subset of
detections associated with each such track.  We assume that a track trajectory
is uniquely determined by the set of detections associated with it and that
some detections may be false positives not associated with any track.

We denote the set of all possible tracks by $\mathcal{P}$ and use $X$ to denote
the detection-track incidence matrix $X \in \{ 0,1\}^{
|\mathcal{D}|\times|\mathcal{P}|}$ where $X_{dp}=1$ if and only if track $p$
visits detection $d$. A solution to the multi-target tracking problem is
denoted by the indicator vector $\pmb{\gamma} \in \{0,1\}^{|\mathcal{P}|}$ where
$\pmb{\gamma}_p=1$ indicates that track $p$ is included in the solution and
$\pmb{\gamma}_p=0$ otherwise.  A collection of tracks specified by $\pmb{\gamma}$ is a
valid solution if and only if each detection is associated with at most one active
track.  Using $\Theta \in \mathbb{R}^{|\mathcal{P}|}$ to denote the costs
associated with tracks where $\Theta_p$ describes the cost of each track $p$,
we express our tracking problem as an integer linear program:
\begin{align}
\label{intobj}
\min_{ \pmb{\gamma} \in \bar{\pmb{\Gamma}}}\Theta^t \pmb{\gamma} 
\quad \quad \quad \text{with}
\quad \quad
\bar{\pmb{\Gamma}} = \{ {\pmb{\gamma}} \in \{ 0,1\}^{|\mathcal{P}|}: \;X {\pmb{\gamma}} \leq 1 \}
\end{align}
Here $\bar{\pmb{\Gamma}}$ is the space of feasible (integer) solutions.  We
note that this is equivalent to finding a maximum-weight set packing where each
set is a collection of detections corresponding to a track and our goal is to
choose a collection of pairwise disjoint sets.  This problem is
NP-hard~\cite{karp} and this formulation faces further difficulties as the size
of the ILP scales exponentially in the number of detections.

\subsection{Decomposing Track Scores over Subtracks}
A classic approach to scoring an individual track is to use a Markov model
that incorporates unary scores associated with individual detections along
with pairwise comparabilities between subsequent detections assigned to a
track. Such an approach provides efficient inference but is limited in its
ability to model higher-order dynamical constraints.  Instead, we consider a
more general scoring function corresponding to a model in which a track is defined by an ordered sequence of subtracks whose scores in turn depend on 
detections across several frames.

Let $\mathcal{S}$ denote a set of subtracks, each of which contains $K$ or
fewer detections where $K$ is a user defined modeling parameter that trades off
inference complexity and modeling power. For a given subtrack $s \in
\mathcal{S}$, let $s_k$ indicate the $k$'th detection in the sequence
$s = \{s_1,\ldots,s_K\}$ ordered by time from earliest (left) to latest (right).  We describe the mapping of subtracks to tracks using $T \in \{ 0,1
\}^{|\mathcal{S}|\times |\mathcal{P}|}$ where $T_{sp}=1$
indicates that track $p$ contains subtrack $s$ as a subsequence. 

We decompose track costs $\Theta$ in terms of the subtrack costs $\theta \in
\mathbb{R}^{|\mathcal{S}|}$ where each subtrack $s$ is associated with cost
$\theta_{s}$ and use $\theta_{0}$ to denote a constant cost associated with
instancing a track.  We define the cost of a track $p$ denoted $\Theta_p$ as:
\begin{align}
\Theta_p=\theta_{0}+\sum_{s \in \mathcal{S} } T_{sp}\theta_s 
\end{align}


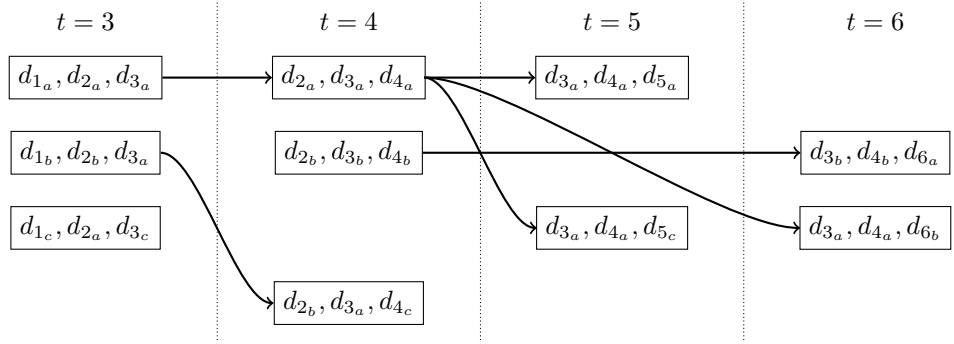
\begin{figure}
\centering
\begin{tikzpicture}
  \node[draw] (node11) at (5,0) {$d_{1_a}, d_{2_a}, d_{3_a}$};
  \node[draw] (node12) at (8.5,0) {$d_{2_a}, d_{3_a}, d_{4_a}$};
  \node[draw] (node13) at (12,0) {$d_{3_a}, d_{4_a}, d_{5_a}$};
  
  \draw[->,thick] (node11) to (node12);
  \draw[->,thick] (node12) to (node13);

  \node[draw] (node21) at (5,-1) {$d_{1_b}, d_{2_b}, d_{3_a}$};
  \node[draw] (node22) at (8.5,-1) {$d_{2_b}, d_{3_b}, d_{4_b}$};
  \node[draw] (node24) at (15.5,-1) {$d_{3_b}, d_{4_b}, d_{6_a}$};

  \draw[->,thick] (node22) to (node24);

  \node[draw] (node31) at (5,-2) {$d_{1_c}, d_{2_a}, d_{3_c}$};
  \node[draw] (node33) at (12,-2) {$d_{3_a}, d_{4_a}, d_{5_c}$};
  \node[draw] (node34) at (15.5,-2) {$d_{3_a}, d_{4_a}, d_{6_b}$};

  \node[draw] (node42) at (8.5,-3) {$d_{2_b}, d_{3_a}, d_{4_c}$};

  \draw[->, thick, in=180, out=0, looseness=0.5] (6,-1) to (7.5,-3);
  \draw[->, thick, in=180, out=0, looseness=0.5] (9.5,0) to (11,-2);
  \draw[->, thick, in=180, out=0, looseness=0.5] (9.5,0) to (14.5,-2);

  \node at (5,0.75) {$t = 3$};
  \node at (8.5,0.75) {$t = 4$};
  \node at (12,0.75) {$t = 5$};
  \node at (15.5,0.75) {$t = 6$};

  \draw[densely dotted] (6.75,1) to (6.75,-3.5);
  \draw[densely dotted] (10.25,1) to (10.25,-3.5);
  \draw[densely dotted] (13.75,1) to (13.75,-3.5);
\end{tikzpicture}
\caption{Diagram of detections:  We use boxes to denote subtracks and use
directed arrows to indicate the valid successors of a given subtrack.  Here we
associate two indicies with a detection.  The first index (numbers) describes the time of
the detection and the second index (letters) describes the particular
observation at that time so $d_{1_a}$ indicates the $a$'th detection at time
$1$. We order the subtracks by the time of their final detection.  Note that a
subtrack may skip some time steps (e.g., $[d_{3_a}d_{4_a}d_{6_b}]$).  This
corresponds to, e.g.,  occlusion where there is no detection observed at
time $5$.} \label{myflowdiag}
\end{figure}

\subsection{LP Relaxation and Column Generation}
\label{lpclique}
We now attack optimization in Eq \ref{intobj} using the well studied tools of
LP relaxations.  We use $\pmb{\Gamma} = \{ \pmb{\gamma} \in [0,1]^{|\mathcal{P}|}: \;X \pmb{\gamma} \leq 1 \}$
to denote a convex relaxation of the constraint set $\bar{\pmb{\Gamma}}$.  The
corresponding relaxed primal and dual problems are written below with dual
variables $\pmb{\lambda} \in \mathbb{R}^{|\mathcal{D}|}$.  
\begin{align}
\min_{ \pmb{\gamma} \in \bar{\pmb{\Gamma}}}\Theta^t \pmb{\gamma} 
\quad \geq \quad \min_{ \pmb{\gamma} \in \pmb{\Gamma}}\Theta^t \pmb{\gamma} 
\quad = \quad \max_{\substack{\pmb{\lambda}\geq 0\\ \Theta+ X^t\pmb{\lambda }\geq 0}} -1^t\pmb{\lambda} 
\label{duallp}
\end{align}

Direct optimization of the dual bound on the right hand side of Eq \ref{duallp}
is still difficult as a consequence of there being one variable in the primal
for every possible track in $\mathcal{P}$ which grows exponentially in the
number of detections.  Hence we employ a column generation approach
\cite{cuttingstock,barnprice} that alternates between solving the optimization
problem with a small active subset of variables, and identifying inactive
variables that may improve the objective then adding these variables to the
active subset.  Identifying such variables corresponds to finding the most
violated constraint (or a set of highly violated constraints including the most
violated) in the dual problem and is computed via combinatorial optimization
(in our case, dynamic programming).

Alg \ref{dualsolvesimple} gives the pseudocode for the column-generation
based optimization.  Here the nascent  subset of  primal variables is denoted
$\hat{\mathcal{P}}$.  We use $\mbox{COLUMN}(\pmb{\lambda})$ to indicate a
subroutine that identifies a group of violated constraints $\dot{\mathcal{P}}$
that includes the most violated given $\pmb{\lambda}$.  Termination occurs
when no more violated constraints exist.

\subsection{Computing $\mbox{COLUMN}(\pmb{\lambda})$ using Dynamic Programming}
\label{colclique}

We now discuss how $\mbox{COLUMN}(\pmb{\lambda})$ is computed
efficiently for our track cost model using dynamic programming.  This is
sometimes referred to as the pricing problem in the column generation literature
\cite{barnprice}.

We specify that a subtrack $s$ may be preceded by a subtrack $\hat{s}$ if and
only if the least recent $K-1$ detections in $s$  correspond to the most recent
$K-1$ detections in $\hat{s}$.  Formally $s_{k-1}=\hat{s}_{k}$ for all k such
that $K\geq k>1$.  We denote the set of valid
subtracks that may precede a subtrack $s$ as $\{\Rightarrow s\}$. This structure
is illustrated graphically in Fig \ref{myflowdiag}. 

To permit the use of tracks with  less than $K$ detections,  subtracks with fewer than $K$ detections are expanded to size $K$ by padded them  with ``no-observation"; denoted  ``$0$", on the left side of the subtrack.  For example consider a subtrack $s=\{ s_1,s_2\}$ where $K=5$.  After padding $s$ with ``no-observation"  detections we write $s$ as  $s=\{ 0,0,0,s_4,s_5\}$ where $s_1$ corresponds to $s_4$ and $s_2$ corresponds to $s_5$. For each ``no-observation" we create a corresponding zero valued $\pmb{\lambda}$ term.     

We use $\ell \in \mathbb{R}^{|\mathcal{S}|}$ to denote the \textit{cost to go}
computed during dynamic programming.  Here $\ell_{s}$ is the cost of the cheapest
track that terminates at subtrack $s$.  Ordering subtracks by the time of last
detection allows efficient computation of $\ell$ using the following dynamic
programming update:
\begin{align}
\label{elldef}
\ell_{s}\leftarrow \theta_s + \pmb{\lambda}_{s_K}+ 
  \min\{ \min_{ \hat{s} \in \{ \Rightarrow s\}}\ell_{\hat{s}},\quad \theta_{0}+\sum_{k=0}^{K-1}\pmb{\lambda}_{s_k} \}
\end{align}

We find it is useful to add not only the minimum cost track (most violated
constraint) to $\mathcal{\hat{P}}$ but also the (most violating) track
terminating at each possible subtrack.  This set of tracks is easy to extract
from the dynamic program since it stores the minimum cost track terminating at
each subtrack.   Only tracks corresponding to violated constraints are added to $\mathcal{\hat{P}}$.   While this over-generation of constraints substantially
increases the number of constraints in the dual, we find that many of these
constraints prove to be useful in the final optimization problem (similar behavior
has been observed in \cite{HPlanarCC}). Additionally, in our implementation
dynamic programming consumes the overwhelming majority of computation time so
adding more columns per iteration yielded faster overall run time.

\subsection{Rounding Fractional Solutions}
\label{round}
We compute upper bounds using a fast principled method that avoids resolving
the LP \cite{lagtrack}.  Observe that each solution of the
LP during the column generation process (Alg \ref{dualsolvesimple}) corresponds
to a (fractional) primal solution in addition to the dual solution (computed
``for free'' by many LP solvers when solving the dual).  We attack rounding a
fractional $\pmb{\gamma}$ via a greedy iterative approach that, at each
iteration, selects the track $p$ with minimum value $\Theta_p\pmb{\gamma}_p$ discounted by the
fractional cost of any tracks that share a detection with $p$ (and hence can no
longer be added to the tracking if $p$ is added).  We write the rounding
procedure in Alg \ref{ubr} using the notation $\mathcal{P}_{\perp p}$ to indicate
the set of tracks in $\mathcal{P}$ that intersect track $p$ (excluding $p$
itself).

\begin{figure}
\begin{tabular}{cc}
\begin{minipage}[t]{7cm}
\null 
 \begin{algorithm}[H]
    \caption{Dual Optimization }
\begin{algorithmic} 
\State $\hat{P} \leftarrow \{ \}$
\Repeat
\State $\pmb{\lambda} \leftarrow \mbox{arg} \max_{\substack{\pmb{\lambda}\geq 0\\ \Theta_{\hat{P}}+ X_{(:,\hat{P})}^t\pmb{\lambda } \geq 0}} -1^t\pmb{\lambda} $
\State $\dot{\mathcal{P}} \leftarrow \mbox{COLUMN}(\pmb{\lambda})$
\State  $\hat{\mathcal{P}}\leftarrow [\hat{\mathcal{P}},\dot{\mathcal{P}}]$

 \Until{ $|\dot{\mathcal{P}}|  =0 $}
\end{algorithmic}
\label{dualsolvesimple}
  \end{algorithm}
\end{minipage}%
&
\begin{minipage}[t]{7cm}
\null
 \begin{algorithm}[H]
    \caption{Upper Bound Rounding}
\begin{algorithmic} 
\While {$\exists p \in \mathcal{P} \quad \mbox{ s.t. } \pmb{\gamma}_p \notin \{ 0,1 \}$}
\State $p^*\leftarrow \mbox{arg}\min_{\substack{p \in \mathcal{P} \\ \pmb{\gamma}_p>0}}\Theta_p\pmb{\gamma}_p -\sum_{\hat{p} \in \mathcal{P}_{\perp p}} \pmb{\gamma}_{\hat{p}}\Theta_{\hat{p}} $ \\
\State $\pmb{\gamma}_{\hat{p}} \leftarrow 0 \quad \forall \hat{p} \in \mathcal{P}_{\perp p^*}$
\State $\pmb{\gamma}_{p^*}\leftarrow 1 $
\EndWhile
\State RETURN $\pmb{\gamma}$
\end{algorithmic}
\label{ubr}
  \end{algorithm}
\end{minipage}\\
\end{tabular}
\caption{(Left):  Algorithm for dual-optimization of a lower bound on the
optimal tracking by column generation where the notation $X_{(:,\hat{\mathcal{P}})}$ denotes
selection of a subset of columns of $X$.  (Right)  We compute upper-bounds on
the optimal tracking using a rounding procedure which greedily selects primal
variables $\pmb{\gamma}$ while removing intersecting tracks.  This same upper
bound procedure is also used during column/row generation in Alg
\ref{dualsolvesimplecyc}.}
\end{figure}

\subsection{Anytime Lower Bounds}
\label{lowerdiscussion}

It is useful in practice to be able to compute a lower bound on the original
objective during the optimization procedure (i.e., prior to adding all the
violated columns to the dual). In Appendix \ref{lowerdiscussionAPP} we show that it is possible to
compute such an anytime lower bound using the following formula.
 \begin{align}
 \label{lowerform}
\min_{ \pmb{\gamma} \in \pmb{\bar{\Gamma}}}\Theta^t \pmb{\gamma} 
  \geq -1^t\pmb{\lambda}+\sum_{d \in \mathcal{D}}\min \{ 0, \min_{\substack{s \in \mathcal{S}\\s_K=d}} \ell_s\} \quad \forall \pmb{\lambda} \geq 0
 \end{align}
The lower bound computed in Eq \ref{lowerform} is maximized at termination of Alg 1 with value equal
to the relaxation in Eq \ref{duallp}.  This is because  no violated constraints in the dual exist at termination and hence $\ell_s \geq 0$ for all $s \in \mathcal{S}$.  Empirically the bound increases as a function of optimization time. 

\section{Tightening the Bound}
\label{betterLP}
Our original LP relaxation only contains constraints for collections of tracks
that share a common detection.  From the view point of maximum-weight
set packing, this includes some cliques of conflicting sets but misses many
others.

\subsection{Fractional solutions from mutually exclusive triplets}
As a concrete example, we consider a case where the LP relaxation Eq
\ref{duallp} provides a loose lower bound which is visualized in Fig
\ref{looseExpFig}.  Consider four tracks $\mathcal{P}=\{p_1,p_2,p_3,p_4\}$ over
three detections $\mathcal{D}=\{ d_1,d_2,d_3\}$ where the first three tracks
each contain two of three detections $\{d_1,d_2\},\{d_1,d_3\},\{d_2,d_3\}$, and
the fourth track contains all three $\{d_1,d_2,d_3\}$.  Suppose the track costs
are given by $\Theta_{p_1}=\Theta_{p_2}=\Theta_{p_3}=-4$ and $\Theta_{p_4}=-5$.
The optimal integer solution sets $\pmb{\gamma}_{p_4}=1$, and has a cost of
$-5$.  However the optimal fractional solution sets
$\pmb{\gamma}_{p_1}=\pmb{\gamma}_{p_2}=\pmb{\gamma}_{p_3}=0.5$;
$\pmb{\gamma}_{p_4}=0$ which has cost $-6$.  Hence the LP relaxation is loose
in this case.  Even worse, rounding the fractional solution results in a
sub-optimal solution. 
 
\subsection{Tightening the Bound over Triplets of Detections}
\label{3fix}

A tighter bound can be motivated by the following observation.  For any set of
three unique detections the number of tracks that pass through two or more
members can be no larger than one.  Thus the following inequality holds for
groups of three unique detections  (which we refer to as triplets)
$d_1,d_2,d_3$.  We use $[...]$ to express the indicator function.  

\begin{align}
\label{lpcorrect3}
\sum_{p \in \mathcal{P}}[X_{pd_1}+X_{pd_2}+X_{pd_3} \geq 2]\pmb{\gamma}_p\leq 1 
\end{align}

We now apply our tighter bound to tracking.  We denote the set of triplets as
$\mathcal{C}$ and index it with $c$.  We denote the subset of $\pmb{\Gamma}$
that satisfy the inequalities in Eq \ref{lpcorrect3} as  $\pmb{\Gamma}^C$ and
define it using a constraint matrix $C \in \{0,1\}^{|\mathcal{C}|\times|\mathcal{P}|}$.
\begin{align}
\pmb{\Gamma}^C: \{ \pmb{\gamma} \in \mathbb{R}^{|\mathcal{P}|}: \pmb{\gamma} \geq 0,\quad X \pmb{\gamma} \leq 1, \quad C \pmb{\gamma} \leq 1 \} 
\end{align}
The constraint matrix has a row for each conflicting triplet specified as follows.  
\begin{align}
\nonumber C_{cp}=[\sum_{d \in c}X_{dp}\geq 2] \quad \forall c \in \mathcal{C}, p \in \mathcal{P}
\end{align}

\section{Optimization over $\pmb{\Gamma}^{C}$}
\label{tightlpopt}
We write tracking as optimization in the primal and dual form below.
\begin{align}
\label{lpobjC}
\min_{\pmb{\gamma} \in \pmb{\Gamma}^C}\Theta^t\pmb{\gamma} 
=\max_{\substack{\pmb{\lambda} \geq 0\\ \pmb{\lambda}^{\mathcal{C}}\geq 0 \\ \Theta+X^t\pmb{\lambda}+ C^t\pmb{\lambda}^{\mathcal{C}} \geq 0}} -1^t\pmb{\lambda}-1^t\pmb{\lambda}^{\mathcal{C}} 
\end{align}

Given that $\mathcal{P}$ and $\mathcal{C}$ are of enormous size we use column
and row generation jointly.  The nascent subsets of $\mathcal{P},\mathcal{C}$
are denoted $\hat{\mathcal{P}}, \hat{\mathcal{C}}$ respectively.
We write column/row generation optimization given subroutines
$\mbox{COLUMN}(\pmb{\lambda},\pmb{\lambda}^{\mathcal{C}})$,
$\mbox{ROW}(\pmb{\gamma})$ that identify a group of violated constraints in
primal and dual including the most violated in each.  Generating rows is done
via exhaustive search and discussed in Section \ref{rowgen}.  Generating
columns is performed using a fast branch and bound procedure where bounding is
done using dynamic programing and is discussed in Section \ref{colbb}.  We
denote the violated columns and rows identified as
$\dot{\mathcal{P}},\dot{\mathcal{C}}$ respectively.  We write the column/row
generation optimization  in Alg \ref{dualsolvesimplecyc}.  Any time upper/lower
bounds are produced using the methods in Sections \ref{round} and
\ref{lowerdiscussion} respectively. Lower-bound computation is modified
from Eq \ref{lowerform} in Section \ref{explower}.    

\begin{tabular}{m{6cm} m{6cm}}
\resizebox{0.45\textwidth}{!}{
\begin{minipage}[t]{7.4cm}
\null 
\begin{tikzpicture}
  \node[draw] (node13) at (7.5,0) {$d_{3_a}$};
  \node[draw] (node14) at (9,0) {$d_{4_a}$};
  \node[draw] (node15) at (10.5,0) {$d_{5_a}$};
  
  \draw[->,thick] (node13) to (node14);
  \draw[->,thick] (node14) to (node15);

  \node[draw,color=white,fill=black] (node21) at (4.5,-1) {$d_{1_b}$};
  \node[draw,color=white,fill=black] (node22) at (6,-1) {$d_{2_b}$};
  \node[draw] (node23) at (7.5,-1) {$d_{3_b}$};
  \node[draw] (node24) at (9,-1) {$d_{4_b}$};
  \node[draw,color=white,fill=black] (node25) at (10.5,-1) {$d_{5_b}$};

  \draw[->,thick,color=blue] (node21) to (node22);
  \draw[->,thick,color=blue] (node22) to (node23);
  \draw[->,thick,color=blue] (node23) to (node24);
  \draw[->,thick,color=blue] (node24) to (node25);

  \draw[->,thick] (node22) to (node13);

  \draw[->,thick] (4.85,-0.9) to (5.65,-0.9);
  \node[draw] (node31) at (4.5,-2) {$d_{1_c}$};
  \node[draw] (node33) at (7.5,-2) {$d_{3_c}$};
  \node[draw] (node34) at (9,-2) {$d_{4_c}$};

  \draw[->,thick,color=red] (node31) to (node22);
  \draw[->,thick,color=red] (node22) to (node33);
  \draw[->,thick,color=red] (node33) to (node34);
  \draw[->,thick,color=red] (node34) to (node25);

  \node[draw] (node42) at (6,-3) {$d_{2_d}$};
  \node[draw] (node43) at (7.5,-3) {$d_{3_d}$};
  \node[draw] (node44) at (9,-3) {$d_{4_d}$};

  \draw[->,thick,color=green] (node42) to (node43);
  \draw[->,thick,color=green] (node43) to (node44);

  \draw[->,thick,color=green] (node21) to (node42);
  \draw[->,thick,color=green] (node44) to (node25);
\end{tikzpicture}
\captionof{figure}{We depict a case where four tracks conflict over a triplet of
detections.  The relevant triplet is $d_{1_b}d_{2_b}d_{5_b}$ with each colored
flow corresponds to a track described in Section \ref{betterLP}. Specifically,
blue flow corresponds to $p_4$ while other colored flows correspond to $p_1$,
$p_2$ and $p_3$ respectively.  Triplets may refer  detections highly separated
in time though this is not depicted in the picture above.}
 \label{looseExpFig}
\end{minipage}
}&
\resizebox{0.475\textwidth}{!}{
\begin{minipage}[t]{7.4cm}
\null
 \begin{algorithm}[H]
 \caption{Column/Row Generation}
\begin{algorithmic} 
\State $\hat{\mathcal{P}} \leftarrow \{ \},\quad \hat{\mathcal{C}} \leftarrow \{ \}$
\Repeat
\State $\max_{\substack{\pmb{\lambda} \geq 0\\ 
\pmb{\lambda}^{\mathcal{C}}\geq 0 \\
\Theta_{\hat{\mathcal{P}}}+X_{(:,\hat{\mathcal{P}})}^t\pmb{\lambda}
+ C_{(\mathcal{\hat{C}},\mathcal{\hat{P}})}^t\pmb{\lambda}^{\mathcal{C}} \geq 0}}
    -1^t\pmb{\lambda}-1^t\pmb{\lambda}^{\mathcal{C}} $
\State Recover $\pmb{\gamma}$ from $\pmb{\lambda}$ (provided by LP solver)
\State $\dot{\mathcal{P}} \leftarrow \mbox{COLUMN}(\pmb{\lambda},\pmb{\lambda}^{\mathcal{C}})$
\State $\dot{\mathcal{C}} \leftarrow \mbox{ROW}(\pmb{\gamma})$
%
\State  $\hat{\mathcal{P}}\leftarrow [\hat{\mathcal{P}},\dot{\mathcal{P}}]$
\State  $\hat{\mathcal{C}}\leftarrow [\hat{\mathcal{C}},\dot{\mathcal{C}}]$
 \Until{ $\dot{\mathcal{P}}=[]$  and $\dot{\mathcal{C}}=[]$ }
\end{algorithmic}
\label{dualsolvesimplecyc}
  \end{algorithm}
\end{minipage}
}
\end{tabular}

\subsection{Row Generation}
\label{rowgen}

Finding the most violated row consists of the following optimization.  
\begin{align}
\max_{c \in \mathcal{C}}\sum_{p \in \mathcal{P}} C_{cp} \pmb{\gamma}_p
\end{align}

Enumerating $\mathcal{C}$ is unnecessary and we generate its rows as needed by considering
only  triplets $c=\{ d_{c_1} d_{c_2} d_{c_3} \}$ such that for each of pair $d_{c_i},d_{c_j}$  there exists a fractional valued track containing both $d_{c_i}$ and $d_{c_j}$.

We find experimentally that adding only the most violated row is efficient, and we only add a row when no violated columns exist.  However in other domains/data sets it may be beneficial to add many violated rows at once and add them even when violated columns exist. 


\subsection{Generating Columns under Triplet Constraints}
\label{colbb}

We denote  the value of the slack corresponding to an arbitrary column $p$
as $V(\Theta,\pmb{\lambda},\pmb{\lambda}^{\mathcal{C}},p)$ and the most
violated as $V^*(\Theta,\pmb{\lambda},\pmb{\lambda}^{\mathcal{C}})$ which we
define below.   
\begin{align}
\label{Vequ}
V(\Theta,\pmb{\lambda},\pmb{\lambda}^{\mathcal{C}},p)&=\Theta_p+\sum_{d \in \mathcal{D}}\pmb{\lambda}_{d} X_{dp} +\sum_{c \in \hat{\mathcal{C}}}\pmb{\lambda}^{\mathcal{C}}_c C_{cp} \\
\nonumber V^*(\Theta,\pmb{\lambda},\pmb{\lambda}^{\mathcal{C}})&=\min_{p \in \mathcal{P}}V(\Theta,\pmb{\lambda},\pmb{\lambda}^{\mathcal{C}},p)
\end{align}
Solving for $V^*(\Theta,\pmb{\lambda},\pmb{\lambda}^{\mathcal{C}})$ can not be
directly attacked using dynamic programming as in Section \ref{colclique}.
However dynamic programming can be applied if we ignore the triplet term
$\sum_{c \in \hat{\mathcal{C}}} \pmb{\lambda}^{\mathcal{C}}_c C_{cp}$,
providing a lower bound.  

This invites a branch and bound (B\&B) approach.  We find B\&B is
very practical because experimentally we observe that the number of non-zero values
in $\pmb{\lambda}^{\mathcal{C}}$ at any given iteration is small ($<5$) for
real problems.  The set of branches in our B\&B tree is denoted
$\mathcal{B}$.  Each branch $b \in \mathcal{B}$ is defined by two sets
$\mathcal{D}_{b+}$ and $\mathcal{D}_{b-}$. These correspond to detections that
must be included on the track and those that must not be included on the track
respectively.  We write the set of all tracks that are consistent with a given
$\mathcal{D}_{b-}$, $\mathcal{D}_{b+}$ or consistent with both
$\mathcal{D}_{b-}$ and $\mathcal{D}_{b+}$ as
$\mathcal{P}_{b-},\mathcal{P}_{b+}$ and $\mathcal{P}_{b\pm}$ respectively.  We
specify the  bounding, branching, and termination operators in Sections
\ref{boundsec},\ref{branchsec}, \ref{termcrit} respectively.  The initial branch $b$ is defined by
$\mathcal{D}_{b+}=\mathcal{D}_{b-}=\{ \}$.

\subsubsection{Bounding  Operation}
\label{boundsec}

Let $V^{b}(\Theta,\pmb{\lambda},\pmb{\lambda}^{\mathcal{C}})$ denote the value
of the most violating slack over columns in $\mathcal{P}_{b\pm}$.  We can compute a lower-bound for this value, denoted $V^{b}_{lb}$ by
independently optimizing the dynamic program and the triplet penalty.
\begin{align}
V^b(\Theta,\pmb{\lambda},\pmb{\lambda}^{\mathcal{C}})&=\min_{p \in \mathcal{P}_{b\pm}} V(\Theta,\pmb{\lambda},\pmb{\lambda}^{\mathcal{C}},p)\\
\nonumber &=\min_{p \in \mathcal{P}_{b\pm}}\Theta_p+\sum_{d \in \mathcal{D}}\pmb{\lambda}_{d}X_{dp} +\sum_{c \in \hat{\mathcal{C}}}\pmb{\lambda}^{\mathcal{C}}_c C_{cp}\\
\nonumber &\geq 
\min_{p \in \mathcal{P}_{b-}}\Theta_p+\sum_{d \in \mathcal{D}}\pmb{\lambda}_{d}X_{dp}
+\min_{p \in \mathcal{P}_{b+}} \sum_{c \in \hat{\mathcal{C}}}\pmb{\lambda}^{\mathcal{C}}_c C_{cp} \\
\nonumber &\geq \min_{p \in \mathcal{P}_{b-}}\Theta_p+\sum_{d \in \mathcal{D}}\pmb{\lambda}_{d}X_{dp} + \sum_{c\in \mathcal{\hat{C}}}\pmb{\lambda}^{\mathcal{C}}_c [\sum_{d \in c} [d \in \mathcal{D}_{b+}]\geq 2] 
\\
\nonumber &=V^b_{lb}(\Theta,\pmb{\lambda},\pmb{\lambda}^{\mathcal{C}})
\end{align}

Observe that dynamic programming can be used to efficiently search over
$\mathcal{P}_{b-}$ to minimize the first term.  For efficiency, subtracks whose
inclusion conflicts with any detection in the required set $\mathcal{D}_{b+}$
can easily be removed before running the dynamic program.

\subsubsection{Branch Operation}
\label{branchsec}

We now consider the branch operation.  We describe an upper bound on $V^{b}(\Theta,\pmb{\lambda},\pmb{\lambda}^{\mathcal{C}})$ as
$V^b_{ub}(\Theta,\pmb{\lambda},\pmb{\lambda}^{\mathcal{C}})$. This is constructed
by adding in the active $\pmb{\lambda}^{\mathcal{C}}$ terms ignored when
constructing $V^b_{lb}(\Theta,\pmb{\lambda},\pmb{\lambda}^{\mathcal{C}})$.  Let
$p_b=\mbox{ arg }\min_{p \in \mathcal{P}_{b-}}\Theta_p+\sum_{d \in
\mathcal{D}}\pmb{\lambda}_{d}X_{dp}$. 

\begin{align}
V^b_{ub}(\Theta,\pmb{\lambda},\pmb{\lambda}^{\mathcal{C}}) &=
V^b_{lb}(\Theta,\pmb{\lambda},\pmb{\lambda}^{\mathcal{C}}) +
\sum_{c\in \mathcal{\hat{C}}}\pmb{\lambda}^{\mathcal{C}}_c C_{cp_b} [\sum_{d \in
c}[d \in \mathcal{D}_{b+}]< 2] \label{upper_bound}
\\
\nonumber &= 
\Theta_{p_b}+\sum_{d \in \mathcal{D}}\pmb{\lambda}_{d}X_{dp_b} + 
\sum_{c\in \mathcal{\hat{C}}}\pmb{\lambda}^{\mathcal{C}}_c [\sum_{d \in c} [d \in \mathcal{D}_{b+}]\geq 2] +
\sum_{c\in \mathcal{\hat{C}}}\pmb{\lambda}^{\mathcal{C}}_c C_{cp_b} [\sum_{d \in c}[d \in \mathcal{D}_{b+}]< 2]
\\
\nonumber &= 
V(\Theta,\pmb{\lambda},\pmb{\lambda}^{\mathcal{C}},p_b) + 
\sum_{c\in \mathcal{\hat{C}}}\pmb{\lambda}^{\mathcal{C}}_c [\sum_{d \in c} [d \in \mathcal{D}_{b+}]\geq 2] - 
\sum_{c\in \mathcal{\hat{C}}}\pmb{\lambda}^{\mathcal{C}}_c C_{cp_b} [\sum_{d \in c}[d \in \mathcal{D}_{b+}]\geq 2] 
\\
\nonumber &= 
V(\Theta,\pmb{\lambda},\pmb{\lambda}^{\mathcal{C}},p_b) + 
\sum_{c\in \mathcal{\hat{C}}}\pmb{\lambda}^{\mathcal{C}}_c (1-C_{cp_b}) [\sum_{d \in c}[d \in \mathcal{D}_{b+}]\geq 2] 
\\
\nonumber &\geq
V(\Theta,\pmb{\lambda},\pmb{\lambda}^{\mathcal{C}},p_b)  
\\
\nonumber &\geq 
V^b(\Theta,\pmb{\lambda},\pmb{\lambda}^{\mathcal{C}})  
\end{align}

Now consider the largest triplet constraint term
$\pmb{\lambda}^{\mathcal{C}}_c$ that is included in
$V^b_{ub}(\Theta,\pmb{\lambda},\pmb{\lambda}^{\mathcal{C}},p_b)$ but not
$V^b_{lb}(\Theta,\pmb{\lambda},\pmb{\lambda}^{\mathcal{C}})$.

\begin{align}
\label{splitEq}
c^*\leftarrow \mbox{arg}\max_{\substack{c\in \mathcal{\hat{C}}}}\pmb{\lambda}^{\mathcal{C}}_c
C_{cp_b} [\sum_{d \in c}[d \in \mathcal{D}_{b+}]< 2]
\end{align}

We create eight new branches for each of the eight different ways of splitting
the detections in the triplet corresponding to $c^*$ between the include ($+$)
and exclude ($-$) sets. In Fig \ref{newsets} we enumerate the splits for a
triplet of detections $c=\{ d_1,d_2,d_3\}$.   In Section \ref{termcrit} we
establish that if $b$ is the lowest cost branch and $\pmb{\lambda}^{C}_{c^*}=0$
then $p_b$ is the track corresponding to the most violated column in
$\mathcal{P}$.  Hence a branch operator is not applied if
$\pmb{\lambda}^{C}_{c^*}=0$.  We refer to a branch $b$ such that
$\pmb{\lambda}^{C}_{c^*}=0$ as terminating.

\begin{figure}
\begin{tabular}{l | l| l ||| l  | l  | l  ||| l  | c | c||| c | c  |c  |c  |c  |}
Child & -& + & Child & - & + & Child              & - & + & Child & - & +    \\
\hline
$D_{b_{1}}$ & $\{ \} $ & $ d_1 d_2 d_3 $& $D_{b_{2}}$ & $ d_1$ & $d_2 d_3$ &$D_{b_{3}}$ & $d_2 $ & $d_1d_3 $& $D_{b_{4}}$ & $d_1 d_2$ & $d_3$  \\
$D_{b_{5}}$ & $d_3  $ & $ d_1,d_2 $& $D_{b_{6}}$ & $ d_1,d_3$ & $ d_2  $ &$D_{b_{7}}$ & $d_2 d_3 $ & $ d_1 $& $D_{b_{8}}$ & $d_1 d_2 d_3$ & $\{ \}$  \\
\end{tabular}
\caption{We enumerate the eight sets each describing one way of partitioning the
three detections $d_1d_2,d_3$ between the include (+) and exclude (-) sets for
the children of branch $b$.  For example, branch $D_{b_4}$ excludes $d_1$ and 
$d_2$ but includes $d_3$ so $D_{b_4-}=[D_{b-}\cup d_1 \cup d_2]$ and the set
$D_{b_4+}=[D_{b+}\cup d_3]$.}
\label{newsets}
\end{figure}


\subsubsection{Establishing Optimality at Termination}
\label{termcrit}
We now establish that B\&B  produces the most violated column at termination. We
do this by proving that the cost of the track corresponding to the lowest cost
branch is both an upper and lower bound on
$V^*(\Theta,\pmb{\lambda},\pmb{\lambda}^{\mathcal{C}})$ if that branch is
terminating.   

Consider branch $b^*$ with corresponding track $p_{b^*}$ such that the following
conditions are true:  (1)  $b^*$ is the lowest cost branch in the B\&B tree; (2)
$b^*$ is terminating.  
We write criteria (1),(2) formally below.
\begin{align}
\label{cirt1}
&(1)\quad V^{b^*}_{lb}(\Theta,\pmb{\lambda},\pmb{\lambda}^{\mathcal{C}})\leq  V^{b}_{lb}(\Theta,\pmb{\lambda},\pmb{\lambda}^{\mathcal{C}}) \quad \forall b\in \mathcal{B}\\
 &(2) \quad 0=\max_{\substack{c\in \mathcal{\hat{C}}}}\pmb{\lambda}^{\mathcal{C}}_c
C_{cp_{b^*}} [\sum_{d \in c}[d \in \mathcal{D}_{b^*+}]< 2] 
\label{cirt2}
\end{align}
We  now establish that  $V(\Theta,\pmb{\lambda},\pmb{\lambda}^{\mathcal{C}},p_{b^*})=V^*(\Theta,\pmb{\lambda},\pmb{\lambda}^{\mathcal{C}})$.
Recall that by definition of B\&B that the lowest value bound in any B\&B tree is a lower bound on the true solution. Therefor Eq \ref{cirt1} implies the following.  
\begin{align}
\label{degboung}
V^*(\Theta,\pmb{\lambda},\pmb{\lambda}^{\mathcal{C}}) \geq V^{b^*}_{lb}(\Theta,\pmb{\lambda},\pmb{\lambda}^{\mathcal{C}})
\end{align}

We now plug Eq \ref{cirt2} into Eq \ref{upper_bound}, and deduce the following:
%

\begin{align}
\label{myeequlist}
V^{b^*}_{lb}(\Theta,\pmb{\lambda},\pmb{\lambda}^{\mathcal{C}})
& \geq  V(\Theta,\pmb{\lambda},\pmb{\lambda}^{\mathcal{C}},p_{b^*})\\
& \geq   V^*(\Theta,\pmb{\lambda},\pmb{\lambda}^{\mathcal{C}}) \nonumber
\end{align}

Observe that Eq \ref{degboung} establishes that  $V^*(\Theta,\pmb{\lambda},\pmb{\lambda}^{\mathcal{C}}) 
\geq V^{b^*}_{lb}(\Theta,\pmb{\lambda},\pmb{\lambda}^{\mathcal{C}})$.
  Therefore all inequalities in Eq \ref{myeequlist} are equalities and hence $p_{b^*}$ is the lowest cost track.  As in Section \ref{colclique} dynamic programming facilitates the addition of many columns per iteration of Alg \ref{dualsolvesimplecyc}.  Any track $p$ produced during calls to dynamic programming during  B\&B such that  $V(\Theta,\pmb{\lambda},\pmb{\lambda}^{\mathcal{C}},p)<0$ should be added to the set $\dot{\mathcal{P}}$ in Alg \ref{dualsolvesimplecyc}.

\section{Experiments}
\label{Exper}

\subsection{Experiments on the Particle Tracking Challenge Data }
\label{experpart}
  
  We applied our algorithm to the data from the Particle Tracking Challenge
  (PTC) simulated data set for high density microtubules (SNR7),which
  consists of two videos (train,test) of 99 images of $512\times 512$
  pixels over time where the  test data set contains $6733$ tracks that cover $71035$
  detections. We take the set of ground truth detections as the set of
  detections and apply our tracking algorithm.

From the set of detections we generate a matrix of possible subtracks of $K$
detections as follows.  Consider a directed graph where the nodes are the set
of all detections.  For each detection $d$ we draw a line from $d$ to each of
its three spatial nearest neighbors in the following frame. The set of all
paths containing $K$ nodes in this graph is the set of subtracks.  In total we
find 1,873,341 subtracks.  We are provided with costs for each subtrack via
logistic regression based on motion features consisting of autocorrelation, and
autocovariance, and other distance features.

%
%
%
%

With $K=3$ and optimized hyper-parameters we reach a Jaccard score of $0.924$
as compared to a baseline of  $0.754$ . We identified $6329$ tracks that are in
the ground truth, missed 404 tracks that are in the ground truth and identified
$118$ tracks that are not in the ground truth.  These results are produced via
providing our output to the benchmarking code associated with
\cite{chenouard14}. In Fig \ref{quantprb}, we applied Alg \ref{dualsolvesimple}
and study tracking performance in terms of accuracy and cost. 


\subsection{Tracking Pedestrians in Video}
\label{natimexp}
\begin{figure*}[th]
\begin{center}
\begin{tabular}{cccc}
\includegraphics[clip,trim=1cm 6cm 1cm 6cm,width=0.3\textwidth]{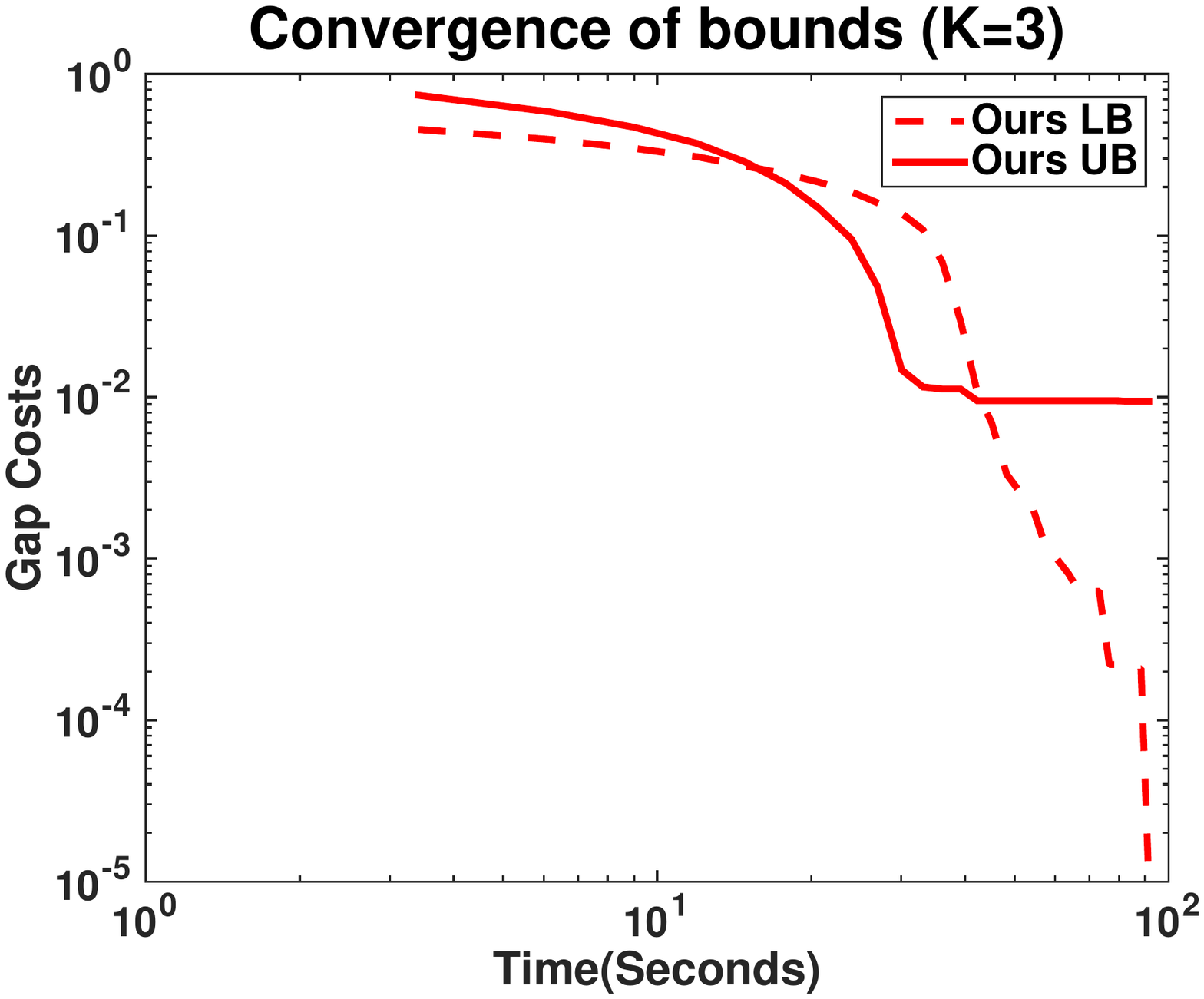}&
\includegraphics[clip,trim=1cm 6cm 1cm 6cm,width=0.3\textwidth]{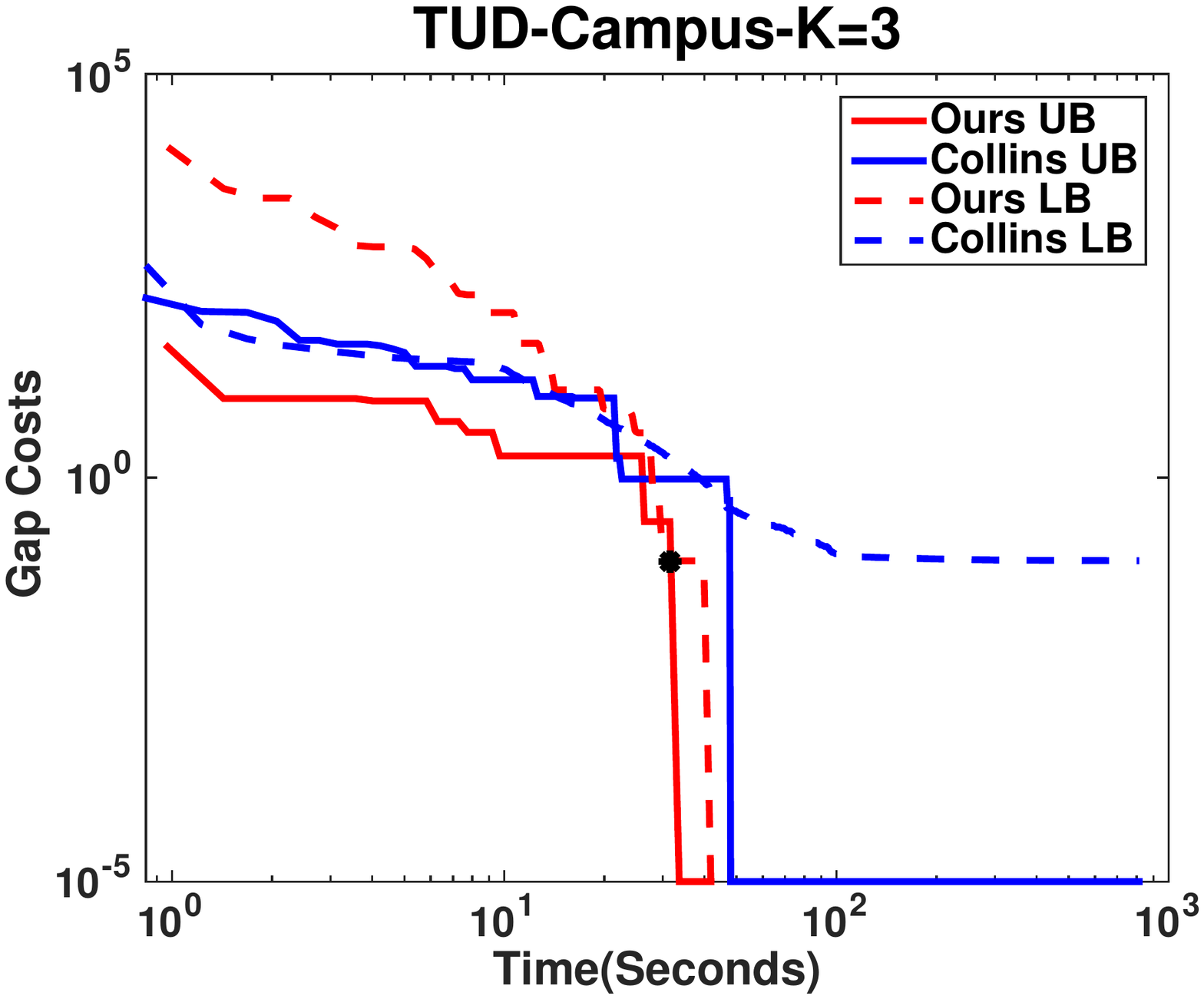}&
\includegraphics[clip,trim=1cm 6cm 1cm 6cm,width=0.3\textwidth]{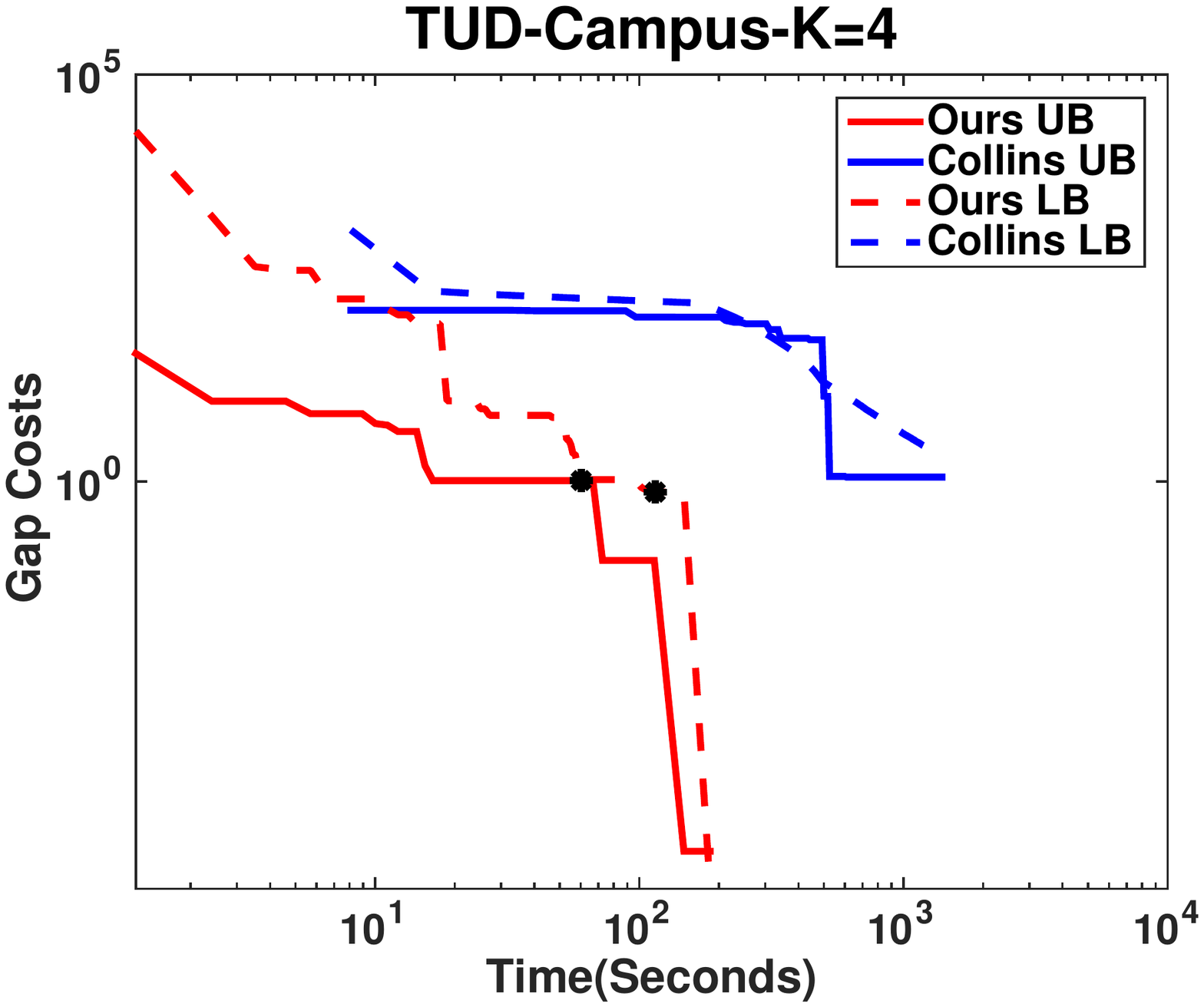}\\
(a) & (b) & (c)
\end{tabular}
\end{center}
\caption{(a): For $K=3$ on Particle Tracking Challenge data, we show the convergence
of the bounds as a function of time in seconds on the test data set. We display
the value of the upper, and lower bounds.  (b) and (c): when training on a
subset of motion features on MOT dataset we get instances with loose bound. For
the two examples we plot the gap (absolute value of the difference) between the bounds and the final lower bound
as a function of time.  We indicate each time that a triplet is added with a black
dot on the lower bound plot.  In all examples the bound of \cite{lagtrack} is
loose and at least one triplet is needed to produce a tight bound which results
in visually  compelling trackings.   We find that additional features for the
structured SVM results in a tighter bound in practice.  
In both comparisons against \cite{lagtrack} our upper- and lower- bounds are
tight at termination.
} 
\label{quantprb}
\end{figure*}

We use a part of MOT 2015 training set~\cite{MOTChallenge2015} to train and
evaluate real-world tracking models. MOT dataset consists of popular pedestrian
benchmark datasets such TUD, ETH and PETS. Specifically we use the learning
framework of~\cite{wang2015learning} with Kalman Filters to train models using
ETH-Sunnyday and TUD-Stadtmitte, and test the models on TUD-Campus sequence. For
detections we use the raw detector output provided by the MOT dataset. We
train the models with varying tracklet length ($K = 2, 3 ,4$) and allow for occlusion up
to three frames. There are altogether 71 frames and 322 detections in the video, 
numbers of subtracks are 1,068, 3,633 and 13,090 for $K = 2, 3 ,4$. For $K=2$
we observe 48.5\% Multiple Object Tracking
Accuracy~\cite{MOTA}, 11 identity switches and  9 track fragments or for short
hand (48.5,11,9).  However when setting $K=3,4$ the performance is (49,10,7),
and (49.9,9,7) which constitutes noticeable improvements over all three
metrics. In Fig \ref{quantprb} we compared the timing/cost performance of our algorithm
with the baseline algorithm of \cite{lagtrack} on problem instances with a
loose lower bound.

\begin{figure*}[th]
\begin{center}
\begin{tabular}{cccc}
\includegraphics[clip,trim=0cm 0cm 0cm 0cm,width=0.25\textwidth]{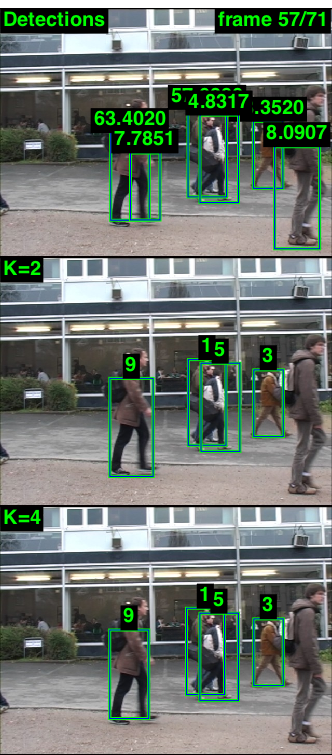}&
\includegraphics[clip,trim=0cm 0cm 0cm 0cm,width=0.25\textwidth]{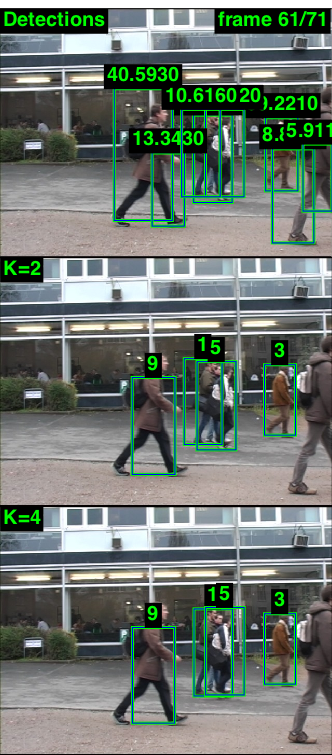}&
\includegraphics[clip,trim=0cm 0cm 0cm 0cm,width=0.25\textwidth]{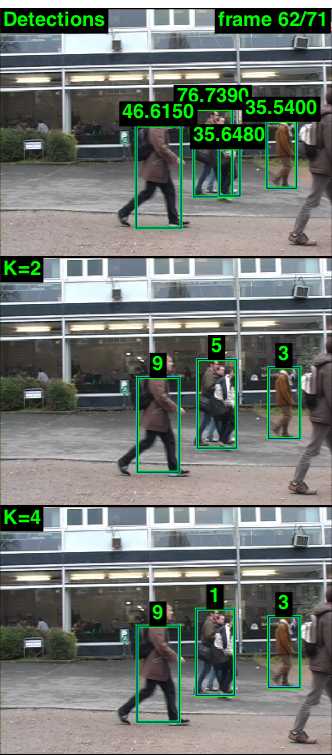}
\end{tabular}
\caption{We illustrate a qualitative example of improvement as a result of
increasing subtrack length. Top row is detector output and associated confidence
provided by~\cite{MOTChallenge2015}. Second row and third row correspond to trackers of subtrack
length $K = 2$ and $K = 4$ respectively. Notice that for $K = 2$ track 1 changes
identity to 5, while with $K = 4$ the identity of track 1 does not change.
Missing detections in tracking results are interpolated linearly and tracks 
are smoothed after interpolation.}
\end{center}
\label{qualres}
\end{figure*}



\section{Conclusions}
\label{conc}

We have introduced a new method for multi-target tracking built on an LP
relaxation of the maximum-weight set packing problem.  Our core contribution is a
column generation approach that exploits dynamic programming to generate a large
number of candidate tracks concurrently.  This yields an efficient algorithm
and provides rigorous bounds
that can be tightened via row generation. We empirically observe that our
algorithm rapidly produces compelling tracking results along with strong
anytime performance relative the baseline Lagrangian relaxation\cite{lagtrack}.

\newpage

\appendix 

\section{Lower Bounds on the Optimal Tracking Cost} 
\subsection{For Relaxation over $\pmb{\Gamma}$}
\label{lowerdiscussionAPP}
We now study lower bounds on the integer programming objective for tracking
which we re-write below using Lagrange multipliers $\pmb{\lambda}$ to 
enforce the constraints defining $\bar{\pmb{\Gamma}}$. 

\begin{align}
\label{intobj1}
\min_{ \pmb{\gamma} \in \bar{\pmb{\Gamma}}}\Theta^t \pmb{\gamma}
=\min_{\substack{\pmb{\gamma} \in \{ 0,1\}^{|\mathcal{P}|}}}
\max_{\pmb{\lambda}\geq 0} \Theta^t \pmb{\gamma}+\pmb{\lambda}^t (X\pmb{\gamma}
-1)
\end{align}

We now add the redundant constraint that no two tracks terminate at the same
detection.  We use $\mathcal{P}^d$ to refer to the set of tracks terminating at
detection $d$.

\begin{align}
\mbox{Eq }\ref{intobj1}=\min_{\substack{\pmb{\gamma} \in \{ 0,1\}^{|\mathcal{P}|}\\\sum_{p \in \mathcal{P}^d}\pmb{\gamma}_p\leq 1 \; \; \forall d \in \mathcal{D} }}\max_{\pmb{\lambda}\geq 0} \Theta^t \pmb{\gamma} +\pmb{\lambda}^t (X\pmb{\gamma} -1)
\end{align}

We now relax the optimization and consider any non-negative $\pmb{\lambda}$.  Recall
that every time dual optimization (during Alg \ref{dualsolvesimple},\ref{dualsolvesimplecyc}) is solved a non-negative $\pmb{\lambda}$
term is produced.  
\begin{align}
\mbox{Eq }\ref{intobj1} &\geq \min_{\substack{\pmb{\gamma} \in \{ 0,1\}^{|\mathcal{P}|}\\ \sum_{p \in \mathcal{P}^d}\pmb{\gamma}_p\leq 1 \; \;  \forall d \in \mathcal{D} }}\Theta^t \pmb{\gamma}
+\pmb{\lambda}^t (X\pmb{\gamma} -1)\\
\nonumber &=-\pmb{\lambda}^t 1+\min_{\substack{\pmb{\gamma} \in \{ 0,1\}^{|\mathcal{P}|}\\ \sum_{p \in \mathcal{P}^d}\pmb{\gamma}_p\leq 1  \; \; \forall d \in \mathcal{D}}}
\nonumber (\Theta^t+\pmb{\lambda}^t X)\pmb{\gamma}\\
\nonumber&= -\pmb{\lambda}^t 1+\sum_{d \in \mathcal{D}} \min \{ 0, \min_{p \in \mathcal{P}^d}\Theta_p +\sum_{d \in \mathcal{D}}X_{dp}\pmb{\lambda}_d \}  \\
 \nonumber &=-\pmb{\lambda}^t 1+\sum_{d \in \mathcal{D}} \min\{ 0,\min_{\substack{s \in \mathcal{S}\\ s_K=d}}\ell_s \}
\end{align} 

At termination of column generation no violated constraints exist so $\ell_s \geq 0 \; \forall s \in \mathcal{S}$ and thus the lower bound has value identical to the LP relaxation over $\pmb{\Gamma}^C$ in Eq \ref{duallp}.

\subsection{For Relaxation over $\pmb{\Gamma}^{C}$}
\label{explower}
 We now consider computing an anytime lower bound on the optimal tracking when  $\pmb{\lambda}^C$ terms are present using 
a similar procedure to that in Section  \ref{lowerdiscussionAPP}.  

\begin{align}
\label{lpobjCA}
\min_{\substack{\pmb{\gamma} \in \{ 0,1\}^{|\mathcal{P}|} \\ \pmb{\gamma}  \in \pmb{\Gamma}^{\mathcal{C}}\\ \sum_{p \in \mathcal{P}^d}\pmb{\gamma}_p\leq 1 }}
\Theta^t\pmb{\gamma} 
&=\min_{\substack{\pmb{\gamma} \in \{ 0,1\}^{|\mathcal{P}|}\\ \sum_{p \in \mathcal{P}^d}\pmb{\gamma}_p\leq 1 \; \;  }} \max_{\substack{\pmb{\lambda}^{\mathcal{C}} \geq 0\\ \pmb{\lambda}\geq 0}} \Theta^t \pmb{\gamma}+\pmb{\lambda}^t(X \pmb{\gamma} - 1)+ \pmb{\lambda}^{\mathcal{C}t}(C \pmb{\gamma}-1) \\
\nonumber&=\min_{\substack{\pmb{\gamma} \in \{ 0,1\}^{|\mathcal{P}|}\\ \sum_{p \in \mathcal{P}^d}\pmb{\gamma}_p\leq 1 \; \;  }}\max_{\substack{\pmb{\lambda}^{\mathcal{C}} \geq 0\\ \pmb{\lambda}\geq 0}}-\pmb{\lambda}^t1-\pmb{\lambda}^{\mathcal{C}t}1+ \Theta^t \pmb{\gamma}+\pmb{\lambda}^t X \pmb{\gamma} + \pmb{\lambda}^{\mathcal{C}t}C \pmb{\gamma}
\end{align}

We now relax the optimization and consider any non-negative $\pmb{\lambda},\pmb{\lambda}^C$.

\begin{align}
\mbox{Eq } \ref{lpobjCA}  
\nonumber&\geq -\pmb{\lambda}^t1-\pmb{\lambda}^{\mathcal{C}t}1+\min_{\substack{\pmb{\gamma} \in \{ 0,1\}^{|\mathcal{P}|}\\ \sum_{p \in \mathcal{P}^d}\pmb{\gamma}_p\leq 1 \; \;  }}   \Theta^t \pmb{\gamma}+\pmb{\lambda}^tX \pmb{\gamma} + \pmb{\lambda}^{\mathcal{C}t}C \pmb{\gamma} \\
 &\geq -\pmb{\lambda}^t1-\pmb{\lambda}^{\mathcal{C}t}1 + \sum_{d \in \mathcal{D}}\min \{ 0,\min_{p \in \mathcal{P}^d}\Theta_p+\sum_{d \in \mathcal{D}}\pmb{\lambda}_{d} X_{dp} +\sum_{c \in \hat{\mathcal{C}}}\pmb{\lambda}^{\mathcal{C}}_c C_{cp} \}
 \label{myLast}
\end{align}
For short hand we define the following quantity $V^{d*}(\Theta,\pmb{\lambda},\pmb{\lambda}^{\mathcal{C}})$ as the lowest cost over tracks terminating at detection $d$.  

\begin{align}
V^{d*}(\Theta,\pmb{\lambda},\pmb{\lambda}^{\mathcal{C}})=\min_{p \in \mathcal{P}^d}\Theta_p+\sum_{d \in \mathcal{D}}\pmb{\lambda}_{d} X_{dp} +\sum_{c \in \hat{\mathcal{C}}}\pmb{\lambda}^{\mathcal{C}}_c C_{cp}
\end{align}

We rewrite the final expression in Eq \ref{myLast} as follows.

\begin{align}
-\pmb{\lambda}^t1-\pmb{\lambda}^{\mathcal{C}t}1+ \sum_{d \in \mathcal{D}}\min \{ 0,\min_{p \in \mathcal{P}^d}\Theta_p+\sum_{d \in \mathcal{D}}\pmb{\lambda}_{d} X_{dp} +\sum_{c \in \hat{\mathcal{C}}}\pmb{\lambda}^{\mathcal{C}}_c C_{cp} \}\\
\nonumber =-\pmb{\lambda}^t1-\pmb{\lambda}^{\mathcal{C}t}1+ \sum_{d \in \mathcal{D}}\min \{ 0,V^{d*}(\Theta,\pmb{\lambda},\pmb{\lambda}^{\mathcal{C}}) \}
\end{align}

We bound $V^{d*}(\Theta,\pmb{\lambda},\pmb{\lambda}^{\mathcal{C}})$ from below in two different
ways. First, we ignore the $\pmb{\lambda}^{\mathcal{C}}$ terms and optimize via dynamic programming producing the following bound.  
$V^{d*}(\Theta,\pmb{\lambda},\pmb{\lambda}^{\mathcal{C}}) \geq  \min_{\substack{s \in \mathcal{S}\\ s_K=d}} \ell_s$.  
%
However we also bound $V^{d*}(\Theta,\pmb{\lambda},\pmb{\lambda}^{\mathcal{C}})$ by the minimizer over all
$d$ hence $V^{d*}(\Theta,\pmb{\lambda},\pmb{\lambda}^{\mathcal{C}}) \geq V^*(\Theta,\pmb{\lambda},\pmb{\lambda}^{\mathcal{C}})$.  Combining the two bounds on $V^{d*}(\Theta,\pmb{\lambda},\pmb{\lambda}^{\mathcal{C}})$ we produce the following bound.  
\begin{align}
\label{finaltoolbound}
V^{d*}(\Theta,\pmb{\lambda},\pmb{\lambda}^{\mathcal{C}})\geq \max \{ V^*(\Theta,\pmb{\lambda},\pmb{\lambda}^{\mathcal{C}}), \min_{\substack{s \in \mathcal{S}\\ s_K=d}} \ell_s \}
\end{align}
We now apply the bound in Eq \ref{finaltoolbound} to produce an anytime computable lower bound on the optimal tracking.  
\begin{align}
\nonumber \min_{\substack{\pmb{\gamma} \in \{ 0,1\}^{|\mathcal{P}|} \\ \pmb{\gamma}  \in \pmb{\Gamma}^{\mathcal{C}}\\ \sum_{p \in \mathcal{P}^d}\pmb{\gamma}_p\leq 1 }}\Theta^t\pmb{\gamma}  
\nonumber & \geq  -\pmb{\lambda}^t1-\pmb{\lambda}^{\mathcal{C}t}1+ \sum_{d \in \mathcal{D}}\min \{ 0,\min_{p \in \mathcal{P}^d}\Theta_p+\sum_{d \in \mathcal{D}}\pmb{\lambda}_{d} X_{dp} +\sum_{c \in \hat{\mathcal{C}}}\pmb{\lambda}^{\mathcal{C}}_c C_{cp} \} \\
\nonumber &=-\pmb{\lambda}^t1-\pmb{\lambda}^{\mathcal{C}t}1+ \sum_{d \in \mathcal{D}}\min \{ 0,V^{d*}(\Theta,\pmb{\lambda},\pmb{\lambda}^{\mathcal{C}}) \}\\
& \geq-\pmb{\lambda}^t1-\pmb{\lambda}^{\mathcal{C}t}1 + \sum_{d \in \mathcal{D}}\min \{ 0,\max\{ V^*(\Theta,\pmb{\lambda},\pmb{\lambda}^{\mathcal{C}}), \min_{\substack{s \in \mathcal{S}\\ s_K=d}} \ell_s \} \}
\end{align}

At termination of column/row generation no violated constraints exist so $V^*(\Theta,\pmb{\lambda},\pmb{\lambda}^{\mathcal{C}})=0$ and thus the lower bound has value identical to the LP relaxation over $\pmb{\Gamma}^C$ in Eq \ref{lpobjC}.

\newpage
\bibliography{example_paper}
\bibliographystyle{ieee}
\end{document}